\documentclass[sigconf]{acmart}
\AtBeginDocument{%
  \providecommand\BibTeX{{%
    \normalfont B\kern-0.5em{\scshape i\kern-0.25em b}\kern-0.8em\TeX}}}

\copyrightyear{2023} 
\acmYear{2023} 
\setcopyright{acmlicensed}\acmConference[FDG 2023]{Foundations of Digital Games 2023}{April 12--14, 2023}{Lisbon, Portugal}
\acmBooktitle{Foundations of Digital Games 2023 (FDG 2023), April 12--14, 2023, Lisbon, Portugal}
\acmPrice{15.00}
\acmDOI{10.1145/3582437.3587207}
\acmISBN{978-1-4503-9855-8/23/04}

\usepackage{xspace}
\usepackage{caption}
\usepackage{subcaption}

\usepackage[clock]{ifsym}

\acmConference[FDG '23]{}{April 11--14,
  2023}{Lisbon, Portugal}
\newcommand*{\mathabxbfamily}{\fontencoding{U}\fontfamily{mathb}\selectfont}
\DeclareFontFamily{U}{mathb}{\hyphenchar\font45}
\DeclareFontShape{U}{mathb}{m}{n}{
      <5> <6> <7> <8> <9> <10> gen * mathb
      <10.95> mathb10 <12> <14.4> <17.28> <20.74> <24.88> mathb12
      }{}

\newcommand*{\Neptune}{{\text{\mathabxbfamily\char"48}}}

\makeatletter
\def\@fnsymbol#1{\ensuremath{
\ifcase#1\or
\Small{\text{\VarTaschenuhr}} \or  
\Neptune \or 
\mathsection \or
\mathparagraph \or
\|\or **
\or
\dagger\dagger
\or \ddagger\ddagger
\else\@ctrerr\fi}}
\makeatother

\renewenvironment{quote}{%
  \list{}{%
    \leftmargin0.5cm   
    \rightmargin\leftmargin
  }
  \item\relax
}
{\endlist}

\begin{document}

\title{“That Darned Sandstorm”: A Study of Procedural Generation through Archaeological Storytelling
}


\author{Florence Smith Nicholls}
\affiliation{%
  \institution{Queen Mary University of London}
  \city{London}
  \country{UK}
}
\email{florence@knivesandpaintbrushes.org }

\authornote{This author should be cited with their full surname "Smith Nicholls" and they/them pronouns.}

\author{Michael Cook}
\affiliation{%
  \institution{King's College London}
  \city{London}
  \country{UK}}
\email{mike@possibiltyspace.org}

\authornote{References to this author may be made using the he/him masculine or they/them singular neutral pronouns.}

\renewcommand{\shortauthors}{Smith Nicholls and Cook}
\newcommand{\nbr}{\textit{Nothing Beside Remains}\xspace}

\begin{abstract}
Procedural content generation has been applied to many domains, especially level design, but the narrative affordances \textit{of} generated game environments are comparatively understudied. In this paper we present our first attempt to study these effects through the lens of what we call a \textit{generative archaeology game} that prompts the player to archaeologically interpret the generated content of the game world. We report on a survey that gathered qualitative and quantitative data on the experiences of 187 participants playing the game \nbr. We provide some preliminary analysis of our intentional attempt to prompt player interpretation, and the unintentional effects of a glitch on the player experience of the game.
\end{abstract}





\maketitle

\section{Introduction}




Procedural content generation is concerned with supporting, modelling and extending creative processes like environmental design. Despite research interest in both procedural narrative and  the generation of game environments, there exists comparatively little research on how procedurally generated environments convey narratives to the player, either intentionally or unintentionally. Procedural generation is often thought of as being unpredictable, hard to control or `random' \cite{nmsgdc}. Yet, evidence from both research and practice suggests that varying kinds of content generator can be controlled, shaped and analysed in multiple ways. Understanding how to provide the same level of variable control for environmental storytelling could empower designers and the development of richer generative systems. We suggest that borrowing methodologies from the field of archaeology, which is traditionally concerned with the interpretation of the \textit{analogue} human material environment, is one way to explore this - and can lead to the creation of what we call \textit{generative archaeology} games, which leverage procedural environmental design as a key part of their game play.

In this paper we present preliminary results from a study conducted in the game \nbr, which procedurally generates a ruined village, implicitly conveying a narrative to the player through environmental design and object descriptions. We prompted participants to play the game and write their interpretation of what happened to the ruined village based on its material remains in order to investigate how their perceptions of a fictional ruin can be altered or affected by the inclusion of both procedural and hand-crafted content.

Our study collected data from 187 participants. They were presented with a survey with questions relating to their experience exploring one of four procedurally generated instances of \nbr, and ther subsequent interpretations. We also collected data on their movements and interactions during play. A fuller analysis of this data will be presented at a future date; in this paper we report on our hypotheses relating to the intentional inclusion of an anachronistic object and the unintentional effects of a glitch on participant emergent storytelling. We also discuss the implications this has for delivering narrative through generated environments, our reflections on conducting the survey and the nature of archaeological storytelling.

The remainder of the paper is organised as follows: in \textit{Background and Related Work} we give the reader an overview of the areas of study that informed this work; in \textit{Methodology} we describe the setup of our study and our aims; in \textit{Preliminary Results} we summarise some of the data gathered and provide analysis of one of our research questions as well as emergent issues that arose during the study; in \textit{Discussion} we contextualise our work and in \textit{Future Work} we outline next steps in terms of both quantitative and qualitative analysis.

\section{Background and Related Work}
\subsection{Generative Archaeology Games}
The term generative archaeology games draws on Cook’s definition of generative forensic games as “a subgenre of information games that challenge the player to understand the output of a generative system” \cite{cook20infogames}. More specifically, generative archaeology games invite the player to explore this output through a process of archaeological interpretation. Archaeologists interpret past activity by recording and analysing material remains. Livingstone et al \cite{daniel2016} consider that environmental storytelling in games is akin to this process of “archaeological storytelling,” creating a narrative through limited material evidence.  This also echoes earlier game studies work, such as Fernández-Vara’s indexical storytelling \cite{fernandez}, and Jenkin’s foundational piece \textit{Game Design as Narrative Architecture}\cite{jenkins}. Both scholars conceive of environmental storytelling as inviting the player to be a detective-a profession which has been likened to archaeology \cite{richardson}.

\subsection{Archaeogaming}
Generative archaeology games are also inspired by the field of archaeogaming; the archaeological study of games in various forms. To date, there has been exciting work in this field concerning the archaeological recording of procedurally generated content, and game development from an archaeological perspective, but limited work \textit{combining} these two together.

Of particular relevance to the themes of our study is Graham’s playthrough of \textit{Minecraft} as an archaeologist, including creative writing responses to the ‘Double Village,’ renowned for a glitch that causes a village to spawn below ground surface \cite{graham2020}. The \textit{No Man’s Sky} Archaeological Survey is perhaps the best example \cite{flick2017} of a project with the aim of applying archaeological methodologies to procedurally generated content in a video game. Reinhard has written about the application of archaeological methods to video games \cite{reinhard2018}, more recently collaborating with Sara Zaia  \cite{reinhard_zaia_2023} on a paper that demonstrates a proof of concept for utilising photogrammetry and GIS to map landscapes in \textit{Fortnite} and \textit{No Man’s Sky}. Linde and Robra \cite{linde} have also done experimental archaeological recording in \textit{Dwarf Fortress}. Though somewhat tangential to archaeogaming, Agent Based Modelling has been used extensively by archaeologists as a way of extrapolating from archaeological data  \cite{graham2020, romanowska}.

In terms of game development from an archaeological perspective, one of the most prolific writers on this topic has been Tara Copplestone; looking at different interactive narrative structures \cite{copplestonedunne} and difficulties in translating archaeological concepts to a video game medium\cite{copplestone2017}. Copplestone’s work is notable for examining archaeological games as a self-reflexive creative endeavour, rather than a purely educational tool\cite{copplestone1}. There has been much valuable work on archaeological storytelling and interactive narrative \cite{value} \cite{vrettakis}, though this is often framed within the context of heritage institutions or with explicitly pedagogical goals in mind. In contrast, our work on generative archaeology games is not mediated by a formal institutional context, exploring how archaeological interpretation is implicitly encouraged through a generative system.


\subsection{\nbr}
\nbr was originally designed in 2018 by the second author for the Procedural Generation Jam, and later written about by them in the context of information games and procedural generation \cite{cook20infogames}. The game is set in the ruins of a settlement, which the player can explore freely, examining objects. While the game has no stated objectives, the original developer notes encourage the player to `discover what happened to the village' they find themselves in \cite{nbritch}.

Each village in \nbr is procedurally generated, using a combination of simple simulation and an embellished rendering step. First, an abstract simulation of a village population is run until it reaches one of three failure states. The simulation models features such as village food supplies, ecosystem health and the prevalence of hostile animals, which affect one another and vary randomly with each simulation tick. Once a failure condition has been hit (such as the collapse of crops) the simulation ends and its state is used to drive generation in the next step.

The village is then generated using an algorithm that is affected by the final state of the simulation, from small-scale details to large structural variations. For example, water sources are placed around the edges of the village, but if the simulation's ecosystem ended in a state that was hot, dry and barren, the water sources will be mostly dried up and smaller in size. The simulation also affects textual descriptions, the distribution of certain object types and the presence or absence of certain buildings. 

Within this algorithmic variation there are several consistent elements. A ruined statue with an inscription inspired by the poem \textit{Ozymandias} is placed near the center of the map in every iteration of the game, and the player always starts at the foot of this statue. Additionally, in every village seed there is always the same large building, usually referred to in documentation as a 'Church', although not explicitly described as such in the game. The contents and details of these two structures vary, but they are guaranteed to exist in every village. Fig. \ref{fig:maps} shows the layout of the two village seeds used in our study, with the statue and church locations marked as S and C respectively.

\subsection{Emergent Narrative}
Emergent narrative in games can be broadly defined as a process by which players construct their own stories through gameplay, often not intended by the original developers. Though emergent narrative is possible in many forms, it has particular relevance with regard to procedural systems due to potential unforeseen outputs and recombination of algorithmically generated content. 

James Ryan’s dissertation \cite{ryan} on \textit{Curating Simulated Storyworlds} argues that a curationist approach is key to producing successful interactive emergent narratives. This framework has been built on by Kreminski, Wardrip-Fruin, and Mateas \cite{kreminski2015}. They discuss using \textit{story sifters} as a way to curate content based on heuristics of what constitutes potentially compelling narrative \cite{kreminski2023}. However, in order to curate for compelling content, one has to understand what makes it interesting in the first place.

One way to sift for interactive narrative content is to use an existing framework from another domain. Lessard and Paré-Chouinard  \cite{lessard}, for example, draw on Georges Polti’s dramatic situations for playwriting. Qualitative and quantitative methods have also been used to evaluate the output of interactive emergent narratives. A quantified analysis of eighty-one playthroughs was applied to \textit{Bad News}, a simulation and performance art piece, in order to inform a story-sifting interface \cite{samuel}. Of particular relevance to our study is Eladhari’s work on game re-tellings as an indicator of emergent narrative quality  \cite{eladhari}. Kreminski et al \cite{kreminski2019} have drawn on this in their qualitative analysis of online retellings of \textit{Civilization VI}, \textit{Stellaris} and \textit{Prom Week}. Our survey of player responses to \nbr could be considered a form of retelling evaluation.

Generally, academic research into procedural content generation has been more concerned with spatial generation than the qualitative generation of narrative \cite{johnson}. In the case of emergent narrative research though, the opposite is true, and the focus is generally on narrative as constructed through text generation. There are some notable exceptions to this trend, one being the GDMC AI settlement challenge, which asks judges to evaluate procedurally generated \textit{Minecraft} settlements based on several criteria, including narrative \cite{gdmc}. There are more examples of work relating to procedural generation and environmental storytelling regarding \textit{embedded narrative} \cite{nielsen2020}, but this work by its very nature is more concerned with the implementation of story vignettes in various recombinations while still maintaining a predetermined narrative, rather than a player-authored one. 

\subsection{Glitches}
The emergent narrative potential of glitches is well established in game studies, as is the compulsion for players to record and share them \cite{meades}. Murnane \cite{murnane} dedicates an entire section of their doctoral thesis on emergent narrative to glitches and the proclivity for players to include them in retellings. Herobrine, a \textit{Minecraft} urban legend, is inspired by visual glitches in the game that often occur at the boundary of it's draw distance \cite{may}. Linking back to archaeogaming, the archaeological potential of glitches as artefacts of a game’s underlying structure has also been explored \cite{smithnicholls,reinhard2018} and provides a further layer of analysis in our discussion of player responses to a glitch in \nbr. 

\section{Methodology}
\subsection{Aims}
This survey is the pilot study for our research into generative archaeology games, using the game \nbr as a case study. The themes of \nbr are particularly appropriate for a generative archaeology game. Part of our motivation in running this work was to better understand how to design player studies of generative archaeology games. In addition to this, we had several specific research questions that we sought to answer:

\begin{itemize}
  \item \textbf{RQ1}: Can players be prompted to interpret what happened to the village in \nbr?
\item \textbf{RQ2}: Do players explicitly refer to the spatial arrangement of different village seeds in \nbr and how does this affect their interpretation of the village?
  \item \textbf{RQ3}: Do players explicitly refer to a deliberately anachronistic object included in some iterations of \nbr and how does this affect their interpretation of the village?
  \item \textbf{RQ4}: Do players confidently assert they can perceive the difference between hand-crafted and procedurally generated content, and if so,  how does this affect their interpretation of the village?
  \item \textbf{RQ5}: Do players express an interest in recording game content in \nbr, and if they do, in what form?
\end{itemize}


As mentioned previously, this paper is primarily concerned with reporting on the design of our experiment, a preliminary analysis of one research question (RQ3), and our reflection on an unexpected glitch that impacted our study. These findings will be useful for both researchers and developers seeking to work with us in this space, and they lay the groundwork for further analysis of the data gathered.

\subsection{Ethical Review}
This study was subject to the King's College London minimal risk self-registration process\footnote{\textbf{Ethical review reference number: MRA-22/23-35186}}. The survey was conducted anonymously through a Google form and no identifying or sensitive information were collected. Participants were provided with an Information Sheet\footnote{\textbf{possibilityspace.org/nbrstudy/information-sheet-nbr-study.pdf}} detailing the research aims of the project and data handling policy. Participants then gave their informed consent before any data was collected. Crucially, participants were informed that data about their movement, duration and interactions within the game would be anonymously recorded and consented to this prior to play.

\subsection{Recruitment}

The only exclusion criteria for the study was that participants had to be over sixteen years of age. The survey was promoted online and open to submissions between the 20th and 25th of January 2023. We were aiming to get at least one hundred responses for a statistically meaningful sample size, and by the 25th we had well-exceeded this threshold. The authors primarily used their own personal social media accounts on Twitter, Mastodon, Discord, TikTok and Facebook in order to recruit participants. Given that the authors are embedded within social media networks predominately comprising peers working in game development, game AI and archaeology, we anticipated there would be a bias towards these groups with pre-existing specialist knowledge in our data. That being said, these groups are also those containing the stakeholders that are most likely to be interested and potentially benefit from the research. We collected demographic information on age and professional relationship to games to provide some context on this.

\subsection {Survey Design}
The survey is split into three sections. The first section requires the participant to consent before continuing. In the second section, participants were given a link to play the game \nbr in a separate window, along with the following instructions:

\begin{quote}
``\nbr is a short game about exploring an abandoned village. Your player character is the `@' symbol. You can use the arrow keys or WASD to move around. Walk into objects to interact with them and read a description. When you are satisfied you have finished exploring, click the "Quit Game" button in the top-right corner of the game, make a note of the code that appears on-screen, and return to this form.''
\end{quote}

A participant is presented with one of four randomly selected versions of \nbr. The bespoke build of the game for the study collected data on player movement, duration of play and interactions so that we could cross-reference which objects a player had encountered with their survey response. This cross-referencing was made possible by the participant entering the personalised code that would appear in a pop-up on screen upon starting and quitting the game. Clicking the `Quit Game' button in the top-right hand corner would automatically send player data to an external server, though data would also be uploaded automatically every fifteen seconds during game play.

In the third section of the survey, participants were asked a series of background questions to contextualise their answers. This included a question as to whether participants had played \nbr before, as it has been available online since 2018 and previous playthroughs could affect player interpretations. Furthermore, respondents were also asked if they played games as a hobby, if they worked in the games industry or considered games to be a part of their academic research. As an attempt to gauge familiarity with the term `procedural generation,' we asked respondents to provide a brief definition if possible. The second part of the third section included five questions relating to their interpretation of \nbr with non-mandatory free text responses, and a sixth non-mandatory question split into two parts which asked respondents what method they would prefer for recording their experience in the game and why. Relevant individual survey questions will be discussed in more detail below. The full list of survey questions can be found online\footnote{\textbf{possibilityspace.org/nbrstudy/questions.pdf}}.

\subsection{Seed Selection}
In order to select seeded conditions for the survey we sampled many at random. We aimed to have seeds with the same simulation outcome -- that is, the same reason for the village to collapse, such as crop failure -- as the aim of this survey was not to gauge if players could accurately distinguish between these different outcomes. We chose the `wildlife attack' simulation outcome as we knew this was likely to produce skeletons and damaged objects in the game world which we hoped would prompt player responses. The effect of different simulation outcomes on player interpretation remains a variable for future study.

\begin{figure*}
    \centering
     \begin{subfigure}[b]{0.48\textwidth}
         \centering
         \includegraphics[width=\textwidth]{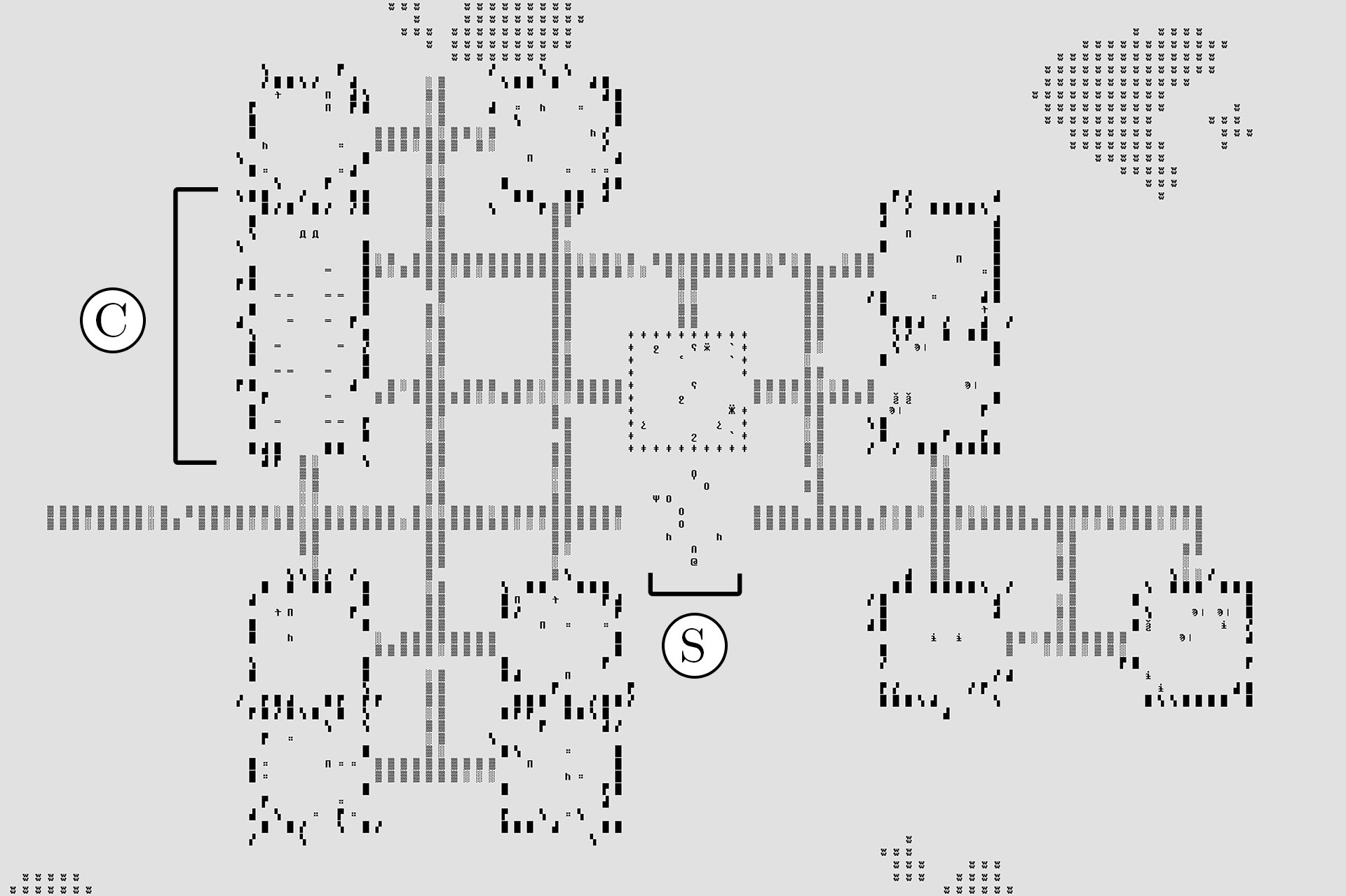}
         \label{fig:map1}
     \end{subfigure}
     \hfill
     \begin{subfigure}[b]{0.48\textwidth}
         \centering
         \includegraphics[width=\textwidth]{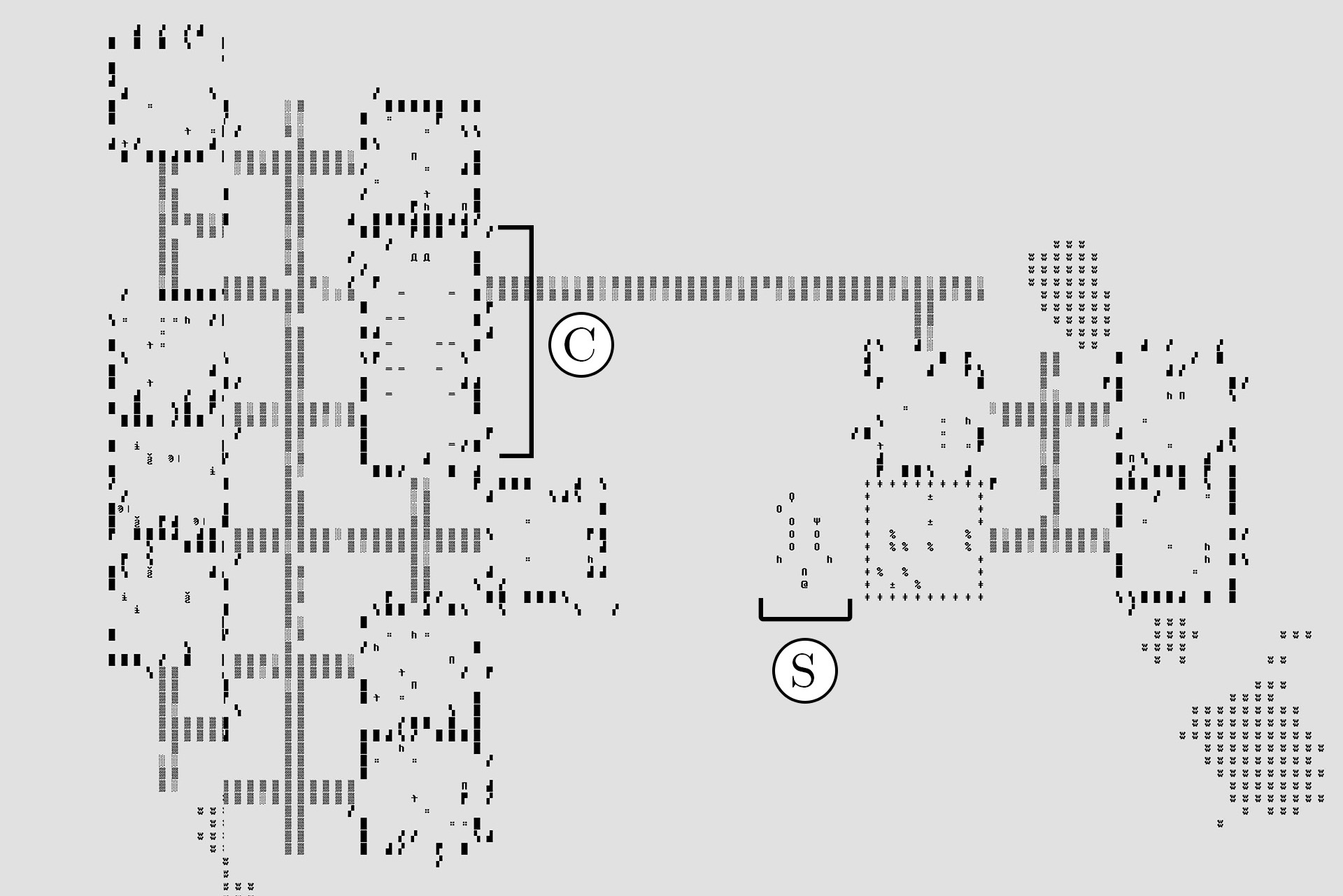}
         \label{fig:map2}
     \end{subfigure}
     \caption{Composite maps of the two seeds used in the study. The starting location and church are marked with the letters S and C respectively.}
     \label{fig:maps}
\end{figure*}

Two seeds were selected for the study with RQ2 in mind (see Figure 1). They were deliberately selected for their divergent layouts -- one has a grid-like structure, while the other is split into two distinct areas connected by a long path. These two seeds represent interesting areas of the generator's expressive range, and we felt would offer two meaningfully different experiences for players. Fig. \ref{fig:maps} shows a zoomed-out view of both villages in their entirety.

\subsection{`Bait' Item}
In order to test RQ3, we came up with the concept of a 'bait' item -- an object with a distinctive description hand-authored to be seemingly anachronistic in the pseudo-ancient setting of \nbr. We chose to include an item with the following description:

\begin{quote}
``A pocket watch made of a delicate treacle coloured metal is embedded in the dirt. There are deep scratches and a stain partly obscuring the face. You can make out part of an inscription: DON.''
\end{quote}

A pocket watch was chosen as a bait item to be distinctive in the game world\footnote{\textbf{The item also implicitly references the Watchmaker analogy, a theological argument for evidence of a designed universe.}}; there are no other mechanical objects in \nbr. The description of the pocket watch was consciously written with partial information and details that could potentially be interpreted as the result of the 'wildlife attack' simulation outcome. It was hand-placed in front of the altar in the church, a building which always appears in every iteration of \nbr. However, in order to test if the inclusion of the object affected player responses, each of the two seeds could also randomly have the item spawn or not. As such, there were four possible versions of the game that a participant could potentially randomly encounter during the survey.

\section{Preliminary Results}
\subsection{Overview}
We collected 187 responses from participants over the course of five days. The average age of participants, submitted as part of the survey, was 35 (st.dev. 9.6 years). Of these respondents, 174 declared they play games regularly (92\%), 36 identified as working in the games industry (19\%) and 35 as working in games research (18\%), with some participants declared as both.

As described earlier, participants were randomly assigned one of two seeds, and then randomly assigned a version of that seed either with or without a bait object. 87 played the first seed, 97 played the second seed, and three respondents could not have their seed determined due to an error in data collection which we resolved shortly after the study launched. 102 received a version of the game with a bait item, while 85 received a version with no bait item. As mentioned previously, we also gathered data on player movements and interactions with the world, resulting in over eight thousand data records, which we are analysing as part of future work.


In this section we report only on preliminary findings with regards to RQ3, which is concerned with player responses to bait item presence in the world. We are currently working on a larger-scale analysis of both quantitative and qualitative data collected, which we describe in the future work section. 

\subsection{The Influence of Bait Items}
We categorised participant responses to question four of the main survey: ``\textit{Pick one object from the game and describe how you think it came to be left where you found it in the village. Why did you choose this object?}''. We manually gathered similar responses together -- for example, participants who chose the feet or head of the ruined statue were all recorded as `statue'. Table \ref{fig:table1} shows the five most-picked objects, broken down into baited and non-baited conditions.

\begin{figure}[t]
\centering
     \begin{subfigure}[b]{0.2\textwidth}
\begin{tabular}{ll}
Item & Count\\\hline
Statue & 15 \\
Chair & 10 \\
Skeleton & 9 \\
Ceramic pieces & 7\\
Altar & 5\\
\end{tabular}
\caption{Non-bait condition}
\end{subfigure}
\hfill
\begin{subfigure}[b]{0.2\textwidth}
\begin{tabular}{ll}
Item & Count\\\hline
Pocketwatch & 20 \\
Statue & 16 \\
Trident & 12 \\
Chair & 6\\
Altar & 4\\
\end{tabular}
\caption{Bait condition}
\end{subfigure}
\caption{The five most commonly-selected items by survey participants to record in the survey.}
\label{fig:table1}
\end{figure}

\begin{figure}[t]
\centering
\begin{tabular}{l|ccc}
Item & Statue & Trident & Pocketwatch\\\hline
On screen & 164 & 163 & 54 \\
Interacted With & 161 & 141 & 51 \\
Recorded & 31 & 11 & 18 \\
\end{tabular}
\caption{A breakdown of three items and how often they were on screen, interacted with, and subsequently chosen as the single item to record.}
\label{fig:table2}
\end{figure}

We can see from the table that in the baited condition the pocket watch was the most frequently picked object, with 20 of the 102 respondents choosing it. In both conditions the statue was very commonly chosen. Notably, the trident -- which is found near the statue at the beginning -- is chosen commonly in the baited condition but less commonly in the non-baited condition (just two times). We are yet to find an explanation for this, however future exploration of play data might shed light on this.

As described earlier, the player always starts in front of the statue. There is exactly one statue in each village, and it stands out as a unique and narratively important structure. By contrast, the pocket watch is located in the church, which is one of a dozen or more buildings, and in both seeds was more than three full screen-spans away from the player's starting location. Thus, its presence as the most chosen object is significant, as it shows that players explored enough to find it, and remembered it later. We theorise that the anachronistic nature of the object helped it stand out, rather than it simply being unique in that world -- other unique items, such as the altar or the statue, were picked less despite their apparent significance. This is reflected in the respondents' interpretations of the pocket watch, as detailed below.


To investigate the difference in object selections, we extracted player interaction and vision data for three objects: the pocket watch bait item, the trident, and the statue. We separated the trident and the statue as the former has a distinctive appearance and was explicitly mentioned many times by participants. For the statue, we combined any interactions or selections with any part of the statue, including its feet, body, head or the pedestal. An object is considered to have been seen if it was rendered on-screen. The data is shown in Fig. \ref{fig:table2}. Note that numbers differ from those shown in Fig. \ref{fig:table1}; due to some issues with data collection we had to exclude some records from the player activity data.

We can see from this breakdown of data that most participants who saw the statue, trident or pocket watch subsequently interacted with it. Given that the statue and trident are seen by players immediately upon starting the game, the fact that 94\% of players who saw the pocket watch then examined it shows a sustained level of interest in this object. We are most interested in how many participants, having interacted with an object, decided to record it in the survey as their one chosen item. Our hypothesis was that the pocket watch was more likely to be selected after having been interacted with, because it stands out as being an anachronistic object in the game world. One possible way to assess this is to use the frequency of participants choosing the statue or trident as a baseline. Like the pocket watch, they are unique items that stand out to the player, both in terms of their placement and textual descriptions. If we take as a baseline the probability of either item being recorded after being interacted with, a simple binomial test suggests that the higher selection rate of the pocket watch is statistically significant (p < 0.05) when compared to either the trident or the statue's probability of being recorded in the survey.

A deeper analysis would be required to have more confidence in these conclusions, however. The data is not entirely independent -- participants who chose the pocket watch had previously seen the trident and statue, and therefore we cannot rule out this affecting their decision-making. We also do not account for other factors such as how centrally an item appeared on the screen, how long the player played for before and after interacting with these items, and other potentially confounding factors. A between-groups analysis with a different experimental setup might provide stronger evidence for our conclusions. Nevertheless, we believe this preliminary analysis strongly suggests that the bait item had a statistically significant impact on participant experience, and gives us a good foundation to build on in future work.

\subsection{Bait Item Interpretations}
As stated above, our hypothesis was that the pocket watch, as a distinctive item that seemed incongruous with the rest of the village, would be picked more frequently by respondents who saw it when prompted to choose an item of interest in the survey (Question 4). This ‘bait’ item was also inspired by \textit{intrusive finds} in the archaeological record – more modern artefacts that intrude on earlier deposits due to later disturbance \cite{mytum}. Players picking up on the object's anachronistic nature shows they are thus applying archaeological interpretation by placing it within the wider context of other material remains in the game world. In a way, the pocket watch functioned as a kind of intentional glitch, a feature to deliberately surprise participants and make them question the nature not only of the village but also potentially the underlying simulation.
Of those responses that mentioned the pocket watch, several explicitly referenced it as seeming to be “out of place”:

\begin{quote}
“There was a pocket watch in a house. There were metal objects in the village, so it is possible they produced the watch too, but the other objects were simple things like bowls. The pocket watch seemed out of place. Maybe another explorer simply dropped it.”
\end{quote}

\begin{quote}
“I found a scratched pocket watch that seemed inconsistent with the technology of the rest of the village which suggests I'm not the first outsider to visit its ruins but there was no hint as to what happened to the outsider.”
\end{quote}

\begin{quote}
“The pocket watch seemed out of place with the rest of the narrative;  let's say that a person exploring this abandoned place left it behind accidentally."
\end{quote}

\begin{quote}
“The pocket watch: I got the idea that I was not the first visitor after the abandoning (?) of the village, as the pocket watch in one of the ruins seemed a markedly more modern object.”
\end{quote}

More than pointing out the potentially anachronistic nature of the pocket watch, some participants went further and questioned if others like them had also visited the abandoned village in \nbr. This is one of the strongest examples of how the bait item prompted emergent archaeological storytelling that considered several phases of activity in the village. This will be followed up on in subsequent qualitative analysis. 

\subsection{``The Sandstorm''}
During the period the survey was open, we discovered that \nbr had a glitch which locked the game if a player moved too far outside of the main village boundary. By the time we discovered this, however, respondents were already incorporating the glitch into their interpretations of the village, so we felt that removing it from the build would be tampering with the experiment. However, the glitch did cause frustration for participants who were unable to continue their playthrough when the game crashed. Though only low stakes, this scenario did present an ethical dilemma in terms of the user experience of the survey.

Normally, it should not be possible to leave the play area in \nbr. The player's position in the world is represented by whole-number co-ordinates, and going beyond the co-ordinates represented by the village grid is not permitted. However, due to a bug in the game's original code, trying to exit the village to the north or south can cause an out-of-bounds array access, rendering the player unable to move and softlocking the game. This is easy for a player to discern because the edge of the map typically has no landmarks or other objects, so there is no visual indicator they are not moving. Particle effects of sand blowing past and sound effects of wind continue to play as other parts of the game are still running. This gives the impression of the desert as empty and endless, no matter which directional button is pressed (even though, in reality, the player has not moved at all).

The so-called ``sandstorm'', unlike the pocket watch, was an actual glitch that led to emergent storytelling. Indeed, some participants even included the sandstorm in their interpretations of the village:
\begin{quote}
“Well, since I tried to find the edge of the map and got caught in a sandstorm I couldn't escape, I'm guessing something similar happened to the village.”
\end{quote}

\begin{quote}
 “Maybe this place has became unlivable because of the sandstorm ? I got caught in the sandstorm myself and got stuck/lost in it. Maybe that happened to them”
\end{quote}

Furthermore, some participants even explicitly stated they believed the sandstorm was a designed feature of the game:

\begin{quote}
“Well, the visual effects of the sand storm might have been procedurally generated, but the sandstorm and its location, probably not.”
\end{quote}

\begin{quote}
“I imagine that the position of that darned sandstorm (the edge of the play area) was hand-written.”
\end{quote}

The sandstorm is an interesting case study for us because in some ways it had a desirable effect by prompting players to come up with interpretations from their experience with the game, however by its very nature it is not a feature that could be reliably reimplemented. Jason Grinblat, a developer on \textit{Caves of Qud} has spoken about the potential of glitches to engage players, especially in games with PCG, and that developers should question if they enhance a game before trying to get rid of them \cite{grinblat}. The “sandstorm” makes a case for this argument. 

\section{Discussion}
Even at this preliminary stage, we can say with some confidence that RQ1 has been answered in that we received over a hundred responses to the survey with participants actively engaged in attempting to interpret the village in \nbr (though, arguably, the open nature of this research question leaves little room for negation). However, more analysis is needed to understand the qualitative nature of these responses and how they relate to the quantitative playthrough data that we collected.

\subsection{Limitations}
Reflecting on the study now that it is complete, there are several aspects in which it could be improved upon in future. Firstly, the inclusion of questions asking participants to define procedural generation and state if they could perceive if content was procedurally generated or not may have primed respondents to perceive content as procedurally generated when they would have otherwise not. The inclusion of a deliberately anachronistic item was also mentioned in the Information Sheet which could have also primed respondents to seek out the object. Conversely, we received informal feedback through social media that the game instructions were not clear and that being explicitly told to interpret the past of the village would be preferable.

Though we collected demographic information on participants' professional relationship with games, we did not collect information regarding their professional relationship with the heritage industry, which would have been valuable contextual information. Other demographic data, such as whether participants' first language was English, could have also been instructive. 

Even with these limitations in mind, we received almost two hundred responses with rich potential for further analysis, which we consider to be a great success considering the experimental nature of this work.



\subsection{Archaeological Storytelling}
Probably the most famous example of a game with procedural emergent narrative is \textit{Dwarf Fortress}. Tarn Adams, one of the game’s programmers, has commented that the ability for dwarfs to make engravings has been a particularly powerful prompt for emergent narrative; players can choose the design of the engraving and where it is located \cite{adams}. Thus, the spatial \textit{context} of engravings is key, not just their textual description – and context is a key component of archaeological storytelling.

Even in the short excerpts of player responses included above, there are examples of participants framing their interpretations of the pocket watch in terms of its location and other known artefacts. In a more specific example, we can arguably see a participant apply the archaeological concept of \textit{assemblage} in their interpretation:

\begin{quote}
“The strange thing was the single three legged chair in the top right house that had a table with three plates set. I would have expected to see another two chairs but it suggests they were moved or destroyed by whatever happened to the village.”
\end{quote}

An archaeological assemblage is a group of artefacts that are associated with each other and were likely used contemporaneously. In this quote, a participant notes how the generated assemblage of objects seemed incomplete – if three plates were set, why was there only one chair? The generative algorithm lacks the contextual understanding to link the placement of the chair and the plates, as they are both put into the world with reference only to the prior simulation and random noise, however even if the result was confusing in this case it prompted player speculation about what created this particular assemblage of objects, leading to the formation of an archaeological narrative. We hope to further explore the potential application of archaeological theory to procedural storytelling as we extend this work, and show how it can be used to inform the design and testing of generative systems.

\subsection{The Glitch as Queer Artefact}
Within game studies, there is precedent for considering that glitches \textit{queer} gameplay experience – they challenge the status quo of a presumed desire for player control and immersion \cite{ruberg}. In terms of procedural content generation, the glitches generative systems produce throw into relief the tension between the marketing of the technique as a way of increasing efficiency, versus the parallel claim that the surprising results they produce are indicative of creative potential \cite{chia}. The ‘sandstorm’ in our survey is a case in point. Some respondents understandably complained about it impeding their gameplay, yet incorporated it in their emergent narratives. 

The existence of “glitch horror” as a genre of internet fiction  \cite{cooley} demonstrates a wider interest in their narrative potential, especially when they subvert expectations. A Roguelike Celebration talk on \textit{Pokémon Glitch} (a randomly corrupted copy of \textit{Pokémon Gold}) has also demonstrated the emergent storytelling potential of game glitches in a roguelike form \cite{rabii}.  If we also consider that glitches themselves constitute archaeological artefacts of a game’s system as opposed to its fictional narrative, we believe our survey results suggest there is potential for playing around with intentional and unintentional glitches to compel players to engage in archaeological interpretation of procedurally generated content. In a sense, some of these players performed a dual archaeological recording within our survey: both of the fictional village, and of the underlying generative system which produced it.

\section{Future Work}

\subsection{Quantitative Analysis}
For each player we have data showing their path through the world during their play session, which objects they interacted with and how often, and what they saw while travelling (that is, what was rendered on-screen). We will analyse the object interactions and cross-reference them with chosen recorded items in the survey. We also aim to connect data of which objects were visible on screen with player path choices, to see if we can establish a connection between the player seeing an object, investigating it, and remembering it. We will also consider clustering players based on their in-game behaviour and analyse possible play styles with a game like \nbr. This might help us understand how players approach open-ended exploration tasks and whether we can design or adapt generators for specific player archetypes.

\subsection{Qualitative Analysis}
The qualitative data provided by the surveys is as rich as the quantitative data. We plan to undertake a grounded theory analysis of player responses to investigate how participants formed interpretations about the village in \nbr. We are particularly interested in how the design of the generative system may implicitly influence player responses, and the potential political implications of this, building on existing research such as Phillips et al’s work on feminist procedural content generation  \cite{phillips}. As stated in the background section, we are also interested in evaluating player responses as a form of gameplay retelling.

\section{Conclusion}
In this paper we reported on a study conducted within the generative archaeology game \nbr, which explored the role of interpretation and inference in a player's understanding of a procedurally generated environment. We described the research which led to this work, and how we designed a study to take advantage of open-ended discovery and environmental storytelling. We also reported on the scope of our data, and showed some preliminary findings which indicate that we were able to successfully shape player experience through the careful inclusion of static content.

Studying open-ended player experience, especially in the context of generated content, raises lots of new challenges for researchers and designers, as well as exciting opportunities for surprising and emergent effects. A bug in the game created a consistent narrative interpretation among survey participants, effectively turning a glitch into a defining characteristic of the game world. We believe that the unpredictable nature of software, like the unpredictable nature of generative algorithms, can be shaped and played with by designers, and that further research will show how understanding emergent narrative as a form of archaeological storytelling is a powerful tool to help us understand these phenomena.

\begin{acks}
The authors wish to thank the reviewers for their helpful feedback. Thanks to Youn\`{e}s Rabii for their formatting of author pronouns in \LaTeX.
\end{acks}

\bibliographystyle{ACM-Reference-Format}
\bibliography{biblio}

\end{document}